\documentclass[a4paper]{spie}  
\usepackage[]{graphicx}

\usepackage{boxedminipage}

\usepackage{fancyheadings}

\title{Extremal optimization for sensor report pre-processing}
\author{Pontus Svenson\\
\bigskip
Department of Data and Information Fusion\\ 
Division of Command and Control Systems,\\
Swedish Defence Research Agency\\
SE 172 90 Stockholm, Sweden \\
\bigskip
{\tt http://www.foi.se/fusion}}
 
\authorinfo{Further author information:\\
E-mail: ponsve@foi.se, Telephone: +46 8 5550 37 32\\  
}

\begin{document}

\thispagestyle{plain}

\maketitle 

\begin{abstract}

We describe the recently introduced extremal
optimization algorithm and apply it to 
target detection and association problems
arising in pre-processing for multi-target
tracking.

Extremal optimization
is based on the concept of self-organized
criticality, and has been used successfully
for a wide variety of hard
combinatorial optimization problems.
It is an approximate local search
algorithm that achieves its success
by utilizing avalanches of local
changes that allow it to explore
a large part of the search space.
It is somewhat similar to genetic algorithms,
but works by
selecting and changing bad chromosomes of a
bit-representation of a candidate solution.
The algorithm is based on processes of self-organization
found in nature. The simplest version of it has
no free parameters, while the most widely used
and most efficient version has one parameter.
For
extreme values of this parameter, the methods
 reduces to 
hill-climbing and random walk searches, respectively. 

Here we consider the problem of 
pre-processing for multiple target
tracking when the number of sensor reports
received is very large and arrives in large
bursts. In this case, it is sometimes
necessary to pre-process reports before sending
them to tracking modules in the fusion system. The
pre-processing step associates reports to known
tracks (or initializes new tracks for reports on
objects that have not been seen before). It could
also be used as a pre-process step before
clustering, {\em e.g.}, in order to test
how many clusters to use.

The pre-processing is done by solving an
approximate version of the original problem.
In this approximation, not all pair-wise conflicts
are calculated. The approximation relies
on knowing how many such pair-wise conflicts
that are necessary to compute. To determine this,
results on phase-transitions occurring when
coloring (or clustering) large random instances
of a particular graph ensemble are used.
\end{abstract}

\section{Introduction}

Tracking is one of the basic
functionalities required in any 
data fusion system. A 
tracker module
takes a number of observations of
an object and uses them to update
our current knowledge on the most
probable position of it. 

Things get more complicated when there are
many targets present, since most
standard trackers~\cite{BlackmanS:ModernTracking}
 rely on knowing exactly from
which object a sensor report stems. In order
to be able to use such modules when there are many
targets present and they are near each-other,
reports must first be pre-processed into
clusters that correspond to the same target.

In this paper, we present two new ideas for
such clustering:
\begin{enumerate}
\item We introduce an approximative way of
calculating the cost-matrix that is needed
for clustering.
\item We use the extremal optimization
method to cluster these approximated
problems.
\end{enumerate}

Clustering is not only used for pre-processing
for trackers. In situations where the flow of reports
is too small to be able to track objects,
clustering of all old and new reports
is used to be able to present a situation-picture
to the user. Clustering is also an
essential part of the aggregation
problem, which deals with reducing the
amount of information presented to users
by aggregating objects into platoons,
companies, and task-forces.

The main reason for using extremal
optimization for the pre-processing proposed
here is that it is a very fast method that
is able to give approximate answers at any
time during execution. Another reason 
for using it is that it is very easy to
extend the method to dynamically add new
reports; the algorithm does not need to
be restarted when a new burst of reports
arrive. 

This pre-processing will become an even more important
problem in the future, as newer, more advanced
sensor systems are used. The presence
of large number of swarming sensors,
for instance, will lead to a much larger number
of reports received simultaneously
than current sensor systems. Still,
the benefit in computation time gained
by using this approximation must be weighted
against the errors that are inevitably introduced
by it.

This paper is outlined in the following way.
Section~\ref{extremal} describes
the background of the extremal optimization
method and explains how to implement
it for clustering. In the following 
section~\ref{preprocess}, the
method for calculating the approximate
cost-matrix is described.
Section~\ref{experiments}
presents the results of some
experiments, while the paper
concludes with a discussion and some
suggestions for future work.

\section{Extremal optimization}
\label{extremal}

Nature has inspired several methods for solving
 optimization problems. 
Among the most prominent examples of such methods
are
simulated annealing~\cite{johnsonaragonmcgeochschevon},
neural nets~({\em e.g.},~\cite{hertzkroghpalmer}),
and genetic algorithms~({\em e.g.},~\cite{mitchell96}).
These methods all rely on encoding a possible
solution to a problem as a bit-string or list
of spin values. A fitness of this string or
spin configuration is then defined
in such a way that an extremum of it corresponds
to the solution of the original problem one
wants to solve.

Recently, the way ant and social insects communicate
to find food have inspired a completely new
set of optimization methods called swarm
intelligence~\cite{bonabeu-book}. A similar method
is extremal optimization, which, like
swarm intelligence, is based on
self-organization. 

Self-organization is the process by which complicated
processes in nature take place without any
central control. An example is how ants find
food --- they do this by communicating information
on where good food sources are by laying
pheromone (smell) paths that attract other ants.
Other examples include the occurrence of
earthquakes and avalanches~\cite{bak:book}. Extremal optimization
is based on models used for simulating
such systems. The Bak-Sneppen model, which is
used to study evolution, consists of $N$ variables
ordered in a chain and given random
fitnesses. In each time-step, the variable
with worst fitness is chosen. It and its neighbors
fitnesses are then updated and replaced with
new random values. Interestingly, this simple
model leads to cascading avalanches of changing
fitnesses that share some of the characteristics
of natural processes, such as extinction of species.
Extremal optimization uses such avalanches of change
to efficiently explore large search landscapes.

Extremal optimization requires defining
a fitness $\lambda_i$ for each variable in the problem.
The fitness can be seen as the local contribution
of variable $i$ to the total fitness of the system.
In each time-step,
the variables are ranked according to their local fitness.

The basic version of extremal optimization then
selects the
variable with the worst fitness in each time-step
and changes it. In contrast to a greedy search,
the variable is always (randomly) changed, even if
this reduces the fitness of the system.
The more advanced $\tau$-EO
instead changes the variable of
rank $k$ with  probability 
proportional to $k^{-\tau}$. 
The value of $\tau$ that 
is best to use depends on the problem at hand.
For detailed descriptions of the algorithm
and its uses for some optimization problems, 
see~\cite{boettcherpercus,boettcher2001,boettcherpercus:cec,boettcherpercusgrigni,informs}.

For a
given problem, the $\tau$ that gives the
best balance between ergodicity and
random walk-behavior should be used.
For extreme values of $\tau$, the method
degenerates:
$\tau= 0$ corresponds to random walk, this is far too ergodic in
that it will wander all over the search space, while
$\tau=\infty$ leads to a greedy search, which quickly gets stuck 
in local minima.

By this process, extremal optimization successively
changes bad variables. In this way, it is actually
more similar to evolution than genetic algorithms.
Extremal optimization stresses the importance
of avoiding bad solutions rather than finding
good ones.

The method has been applied with great success
to a wide variety of problems, including
graph partitioning~\cite{boettcher:pregraph,boettcher}
and coloring~\cite{boettcher:col},
image alignment~\cite{image}
and studies of spin glasses~\cite{bethe,stiffness}.
The method doesn't seem to work well
for problems with a large number of connections
between the variables.

The original extremal optimization algorithm
achieves its emergent behavior of finding
a good solution without any fine-tuning. When
using the $\tau$-extension, it is necessary
to find a good enough value of $\tau$ to
use, but there are some general guide-lines
for this.

The method of updating the worst variable causes
large fluctuations in the solution space. It is
by utilizing these fluctuations to quickly search
large areas of the energy landscape that
the method gains its efficiency.

Experimentally, runs of the extremal optimization
often follow the same general behavior. First,
there is a surprisingly quick relaxation of the
system to a good solution. This is then followed
by a long period of slow improvements. The results
obtained in the experiments described in
section~\ref{experiments} also follow this pattern.

\begin{figure}
\begin{boxedminipage}{\columnwidth}
{\bf Extremal optimization}
\begin{enumerate}
\item Initialize variables randomly
\item While time $<$ maximum time
\begin{enumerate}
\item Calculate local fitness $\lambda_i$ for each variable $i$
\item Sort $\lambda_i$
\item Select $k$'th largest $\lambda_i$ with probability $p_k \sim k^{-\tau}$ ($\tau$-EO)
\item OR select largest $\lambda_i$ (standard EO)
\item Change selected variable, regardless of how this affects cost
\item If new configuration has lowest cost so far
\begin{enumerate}
\item Store current configuration as best
\end{enumerate}
\end{enumerate}
\item Output best configuration as answer
\end{enumerate}
\end{boxedminipage}
\caption{Pseudo-code for the extremal optimization clustering algorithm.}
\label{pseudo}
\end{figure}

To use the method for clustering, we follow
the following simple steps. Using $x_i$ to denote
report $i$, write the
cost function as
a sum of pair-wise conflicts
\begin{equation}
C = \sum_{i<j} C(i,j) \delta_{x_i}^{x_j} .
\end{equation}
The local fitness $\lambda_i$ is easily seen
to be
\begin{equation}
\lambda_i = \sum_{j\neq i} C(i,j) \delta_{x_i}^{x_j} .
\end{equation}

Pseudo-code for the algorithm is shown in
figure~\ref{pseudo}.

Implementing the algorithm requires 
power-law distributed random numbers. This
is best obtained using pre-computation of two lists
of numbers. One list contains powers
\begin{equation}
a_k = k^{-\tau}  ,
\end{equation}
while the other is the cumulative sum of $a_k$
\begin{equation}
b_n = \sum_{k\leq n} a_k .
\end{equation}
The list $b_n$ is needed in order to be able
to handle dynamically changing number of
reports to cluster. The power-law
distribution when there are $n$
reports is now given by
\begin{equation}
p_k^n = \frac{a_k}{b_n} .
\end{equation}

Selecting the appropriate $k$ is done by comparing
successive terms with a uniformly distributed random
number in the standard way to determine
in which interval of $p_k^n$ it fell.

A way of speeding up the extremal optimization 
method is to use a heap instead of
a sorted list for storing the $\lambda_i$. A
heap is a balanced binary tree where each node has a higher value than
each of its children, but there is no requirement that
the children are ordered. Since insertions and deletions
in a heap can be done in logarithmic time, a factor
$n$ is gained in speed. For some problems,
it has been shown that the error introduced by
not sorting the $\lambda_i$ completely
is negligible~\cite{boettcherpercus}. In the implementation
used here, this optimization was not used in order
to be able to be able to test the correctness of
using the approximate cost matrix $c_{ij}$
without introducing possible errors from using a heap.

\section{Clustering as pre-processing for tracking}
\label{preprocess}

The general clustering
or association problem can be formulated as
a minimization problem. Introducing the notation
$x_i=a$ when report $i$ is placed in cluster $a$,
this can be written as
\begin{equation}
\min_{\{x_i\}} C(\{x_i\}) ,
\label{eqfull}
\end{equation}
where $C$ denotes the cost of a configuration.
The cost includes terms that give the cost of
placing reports together, and also the cost of
not placing reports together. 

Clustering problems are often solved
in an approximate way by ignoring
multi-variable interactions and focusing
on pair-wise conflicts. For an example
of how to use Dempster-Shafer
theory to derive such an approximation,
see~\cite{johan}. The approximation
of including only pair-wise conflicts leads to
the expression
\begin{equation}
C({x_i}) \approx \sum_{i<j} C(i,j) \delta_{x_i}^{x_j} .
\label{eq2}
\end{equation} 
It would in principle be possible to include
also higher-order terms in this equation;
for instance, a term giving the cost of placing
three given spins in the same (or different!)
clusters could be included.
Using equation~\ref{eq2} makes it possible
to solve the clustering problem by mapping
it onto a spin model and using any of the
many optimization methods devised for them.
It is also computationally much less demanding
than using the full equation~\ref{eqfull},
since only $N^2$ terms in the $C(i,j)$ matrix
need to be calculated.

For some applications, however, even
these $N^2$ conflicts might require too much
processing power to calculate. This
is the case when a large number of
reports arrive in sudden bursts.
If the reports are to be used to track
many objects, they first need to be assigned
to the correct tracker. (Note: we assume
here that the trackers used are single-target
trackers. If one uses multi-target trackers
based on random sets or probability
hypothesis density, no such association
needs to be done. Ideas for such
trackers have been described in,
among others,~\cite{sidenbladhFUSION03,mahler00}
)

Another motivation for trying to find
an alternative to calculating all $N^2$
conflicts comes from the increasing importance
of distributed architectures for fusion systems.
If the reports are distributed across a large
number of nodes, we want to minimize the amount
of data that needs to be transmitted across
the network. If the clustering is also done
using distributed computing, we need to calculate
as few pair-wise conflicts as possible.

In this paper, we present an alternative to
clustering such bursts of report using the
full cost function or the pair-wise
cost-matrix introduced in equation~\ref{eq2}.

The method is based on the observation that
the clustering problem must have a solution,
and that for many problems there is a sharp
threshold in solvability
when some parameter is varied~\cite{hogghubermanwilliams}.
For the problems considered in this paper, the relevant
phase transition to look at is the one occurring
for the graph coloring (or clustering) problem.
For random graphs with average degree $\gamma$,
the solvability of a randomly chosen instance
is here determined by comparing $\gamma$ with
a critical parameter $\gamma_c$: for 
$\gamma < \gamma_c$, almost all graphs are
colorable ({\em i.e.}, they can be clustered
without cost), while for larger values
of $\gamma$, the probability that such a
clustering exists vanishes. 
For more information on this phase-transition,
the reader is referred
to~\cite{hogghubermanwilliams}. Information on analytical
ways of determining $\gamma_c$ can be found
in~\cite{achlioptas:thesis}, while numerical
values are given in~\cite{nprelax}. Note that
the value of $\gamma_c$ depends on the number
of clusters $k$; for large values
of $k$, various approximations can be used.

Given a set ${x_i}$ of $N$ reports, we therefore
propose
that not all pair-conflicts are calculated. Instead,
we randomly select $M=\frac{1}{2} \gamma N$
edges and calculate the conflict for these. This
gives a sparse matrix $c_{ij}$, that is
a subset of the full $C(i,j)$ matrix:
whenever $c_{ij}$ is non-zero, it is equal to 
$(C(i,j)$, but there might be entries
in $C(i,j)$ that are not represented in $c_{ij}$.

The matrix $c_{ij}$ is thus a member of the
$\mathcal{G}(N,M)$ ensemble of random
graphs, but the exact values of its edges
are determined by the full cost matrix $C(i,j)$.

We stress again that the purpose of doing this
is to avoid costly computations. If it is
cheap to compute all entries in $C(i,j)$,
this should of course be done. If, however,
each of these computations involve sophisticated
processing such as the calculation of Dempster-Shafer
conflicts of complicated belief functions, then
a factor $\frac{N}{\gamma}$ in computation
time is gained by only calculating $c_{ij}$.

The approximate matrix $c_{ij}$ is then used for
clustering the reports. The solution
that is returned is an approximate solution also
of the complete problem only if $c_{ij}$
faithfully captures all the important structure
of $C(i,j)$. In order for it to do this,
we want to use a $\gamma$ that is as close to
the phase-transition as possible. Problems
near the phase-transitions are the maximally
constrained problems, which lends credibility
to the correctness of the approximation $c_{ij}$.

\section{Experiments}
\label{experiments}

The experiments performed all use a similar scenario.
There are 3 enemy objects that are seen
simultaneously by a large number of sensors.
The sensors involved could be for instance
ground-sensor networks~\cite{iam} or
packs of swarming UAVs~\cite{gaudiano03a}.

In the figures presented here, bursts of
100 reports are pre-processed for further transmission
to tracking modules. Since $\gamma_c(3)\approx 4.6$,
we present results for $\gamma=3$, 4, and 5.
We ran several different simulations for each parameter.
Some of the figures show results of a single sample,
while others show averaged results. 

Two averages were performed.
First, a number of different sets of sensor
reports were generated.
Second, a number of different approximate cost-matrices
$c_{ij}$ were calculated given a set of sensor
reports. Changing the number of averages did not affect the
results.

Each of the figures shows two curves as functions
of time: the cost/fitness
of the current solution and the best cost
obtained so far.

For each scenario, a true cost matrix $C(i,j)$ was determined
and used for calculating $c_{ij}$ and for comparing the
results of our method with the true solution. A real
implementation would of course not calculate $C(i,j)$;
it was used here only to test our method.

\begin{figure}
\centering
\includegraphics[width=.50 \columnwidth]{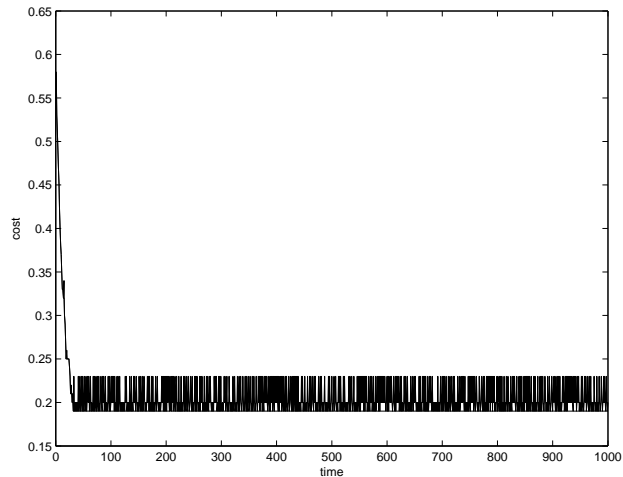}
\vspace{2mm}
\caption{Evolution of best and current cost as a function of time
for standard-EO for a $\gamma=3$-run.}
\label{g3eo}
\end{figure}

\begin{figure}[!t]
\centering
\includegraphics[width=.50 \columnwidth]{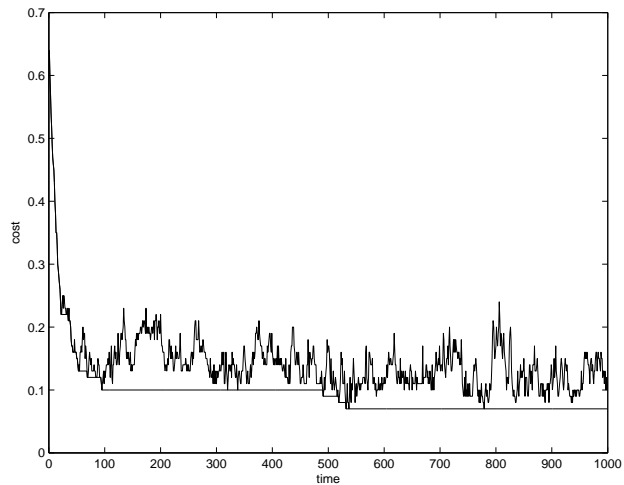}
\vspace{2mm}
\caption{Evolution of best and current cost as a function of time
for $\tau=1.5$-EO for a $\gamma=3$-run.}
\label{g3t15}
\end{figure}

For $\gamma=3$, below the phase transition
figure~\ref{g3eo} and~\ref{g3t15} show the difference between the standard
extremal optimization algorithm and $\tau$-EO. The value
of $\tau$ to use was determined using some experimentation;
it depends on the structure of the problem. It can be seen quite clearly
that $\tau$-EO is better. The figures show the result of
a typical run, with no averages being done.

\begin{figure}
\centering
\includegraphics[width=.50 \columnwidth]{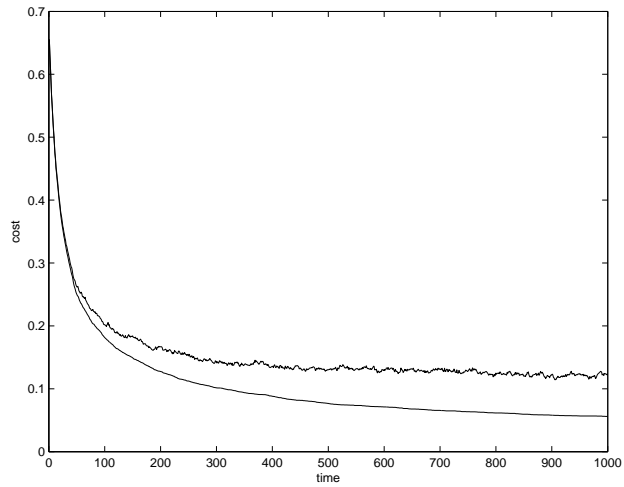}
\vspace{2mm}
\caption{Evolution of best and current cost as a function of time
for $\tau=1.5$-EO for an average over 10 different $\gamma=3$-runs.}
\label{g3t15ave}
\end{figure}

Figure~\ref{g3t15ave} shows the behavior 
for $\gamma=3$ when averaged over
10 different problems. Here, for each problem the algorithm
has also been restarted 10 times using different
approximative cost-matrices.

\begin{figure}
\centering
\includegraphics[width=.50 \columnwidth]{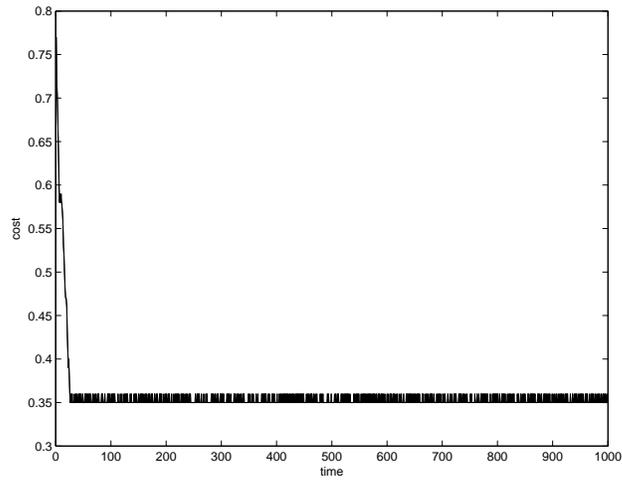}
\vspace{2mm}
\caption{Evolution of best and current cost as a function of time
for standard-EO for a $\gamma=4$-run.}
\label{g4eo}
\end{figure}

What happens when increasing $\gamma$ can be clearly seen by
comparing figures~\ref{g3eo} and \ref{g4eo}, which both
show one run with the standard extremal optimization algorithm
but for  $\gamma=3$ and 4, respectively.

\begin{figure}[!t]
\centering
\includegraphics[width=.50 \columnwidth]{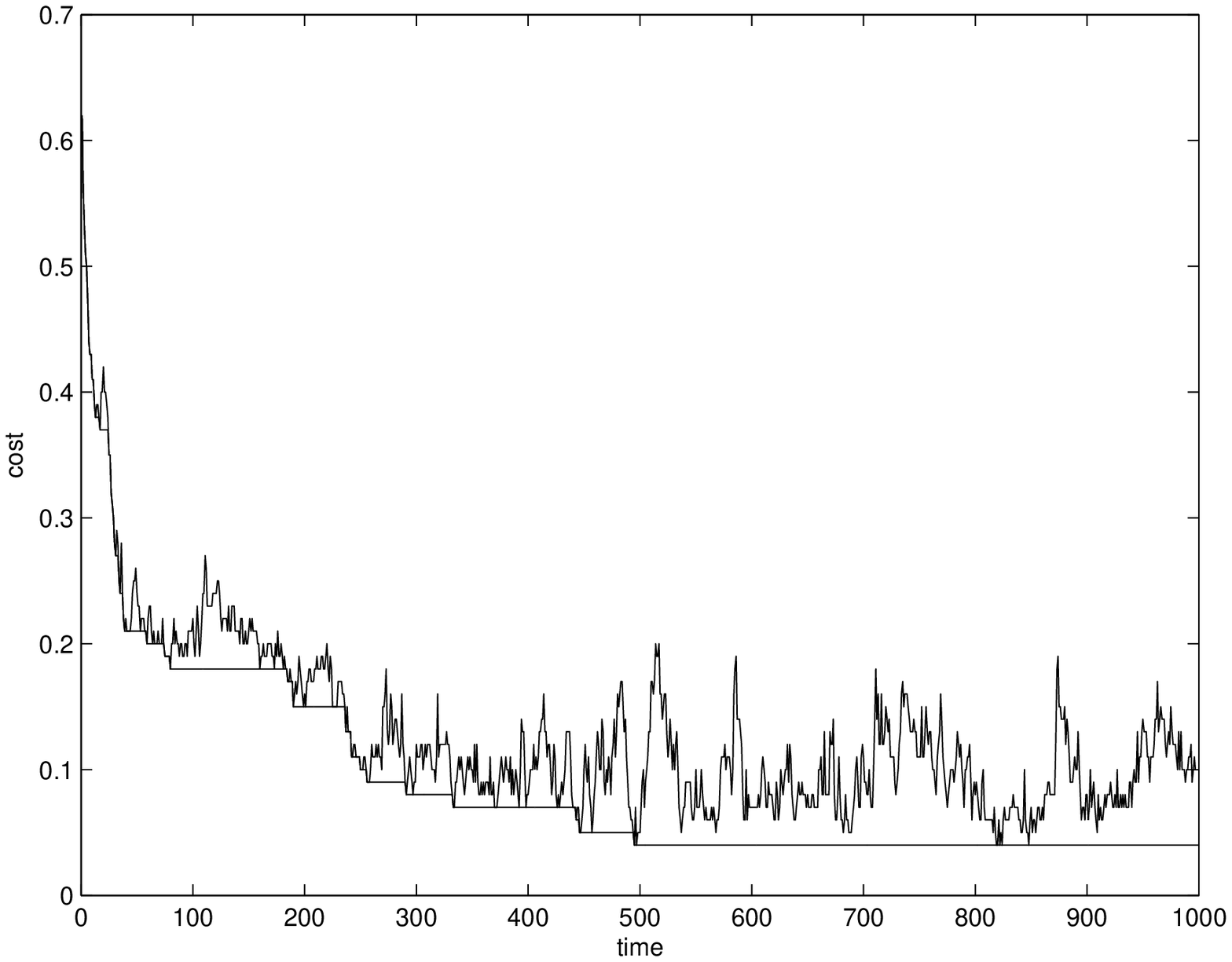}
\vspace{2mm}
\caption{Evolution of best and current cost as a function of time
for $\tau=1.5$-EO for a $\gamma=4$-run.}
\label{g4t15}
\end{figure}

\begin{figure}[!t]
\centering
\includegraphics[width=.50 \columnwidth]{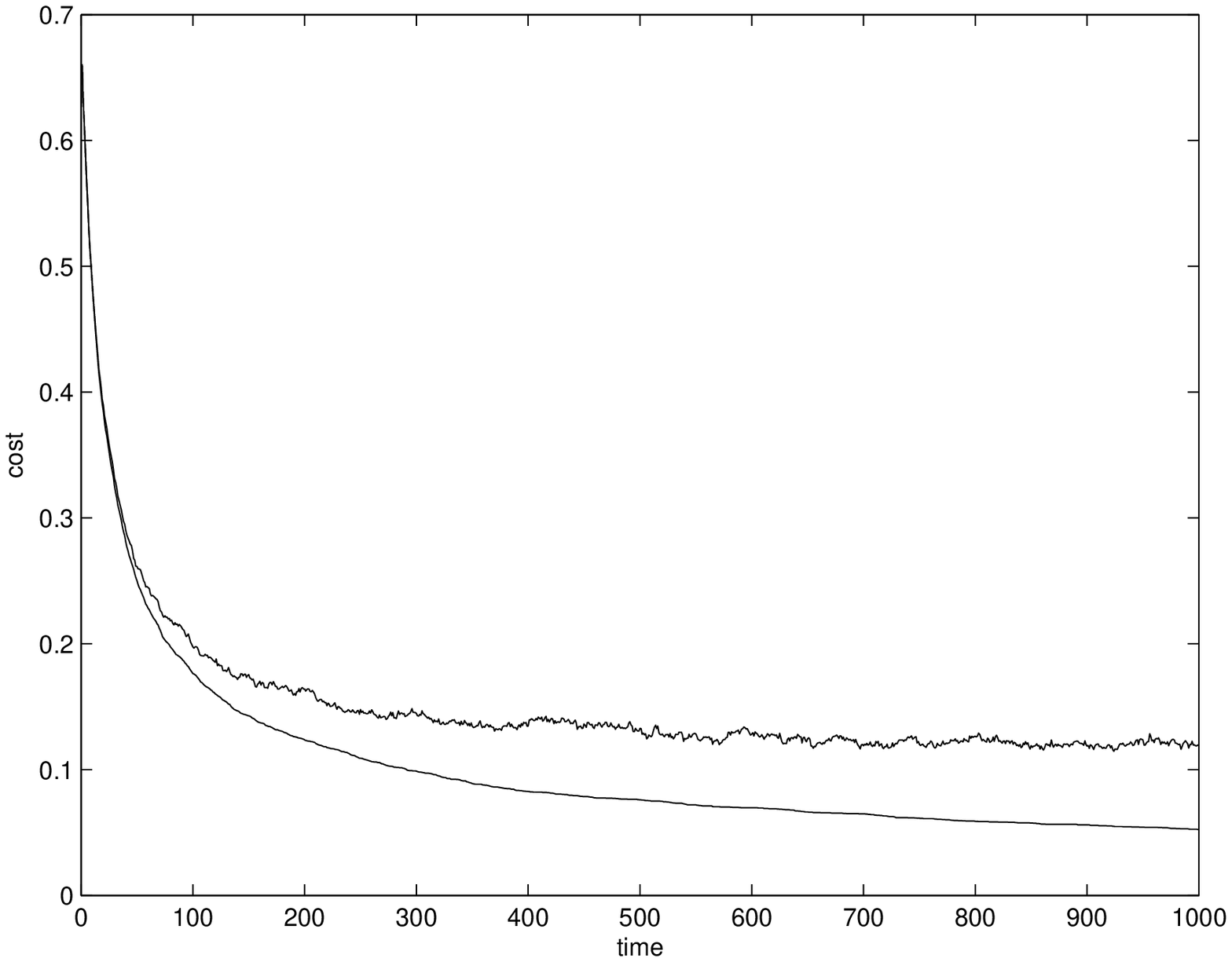}
\vspace{2mm}
\caption{Evolution of best and current cost as a function of time
for $\tau=1.5$-EO for an average over 10 $\gamma=4$-runs.}
\label{g4t15ave}
\end{figure}

A single run with $\tau$-EO for $\gamma=4$ and $\tau=1.5$ is shown in
figure~\ref{g4t15}, while the results
for averages over 10 and 10 graphs are shown
in figure~\ref{g4t15ave}

Now we turn to results for $\gamma=5$, that is
above the phase transition. Here
the approximate cost matrices should
give problems that are not solvable.

\begin{figure}[!t]
\centering
\includegraphics[width=.50 \columnwidth]{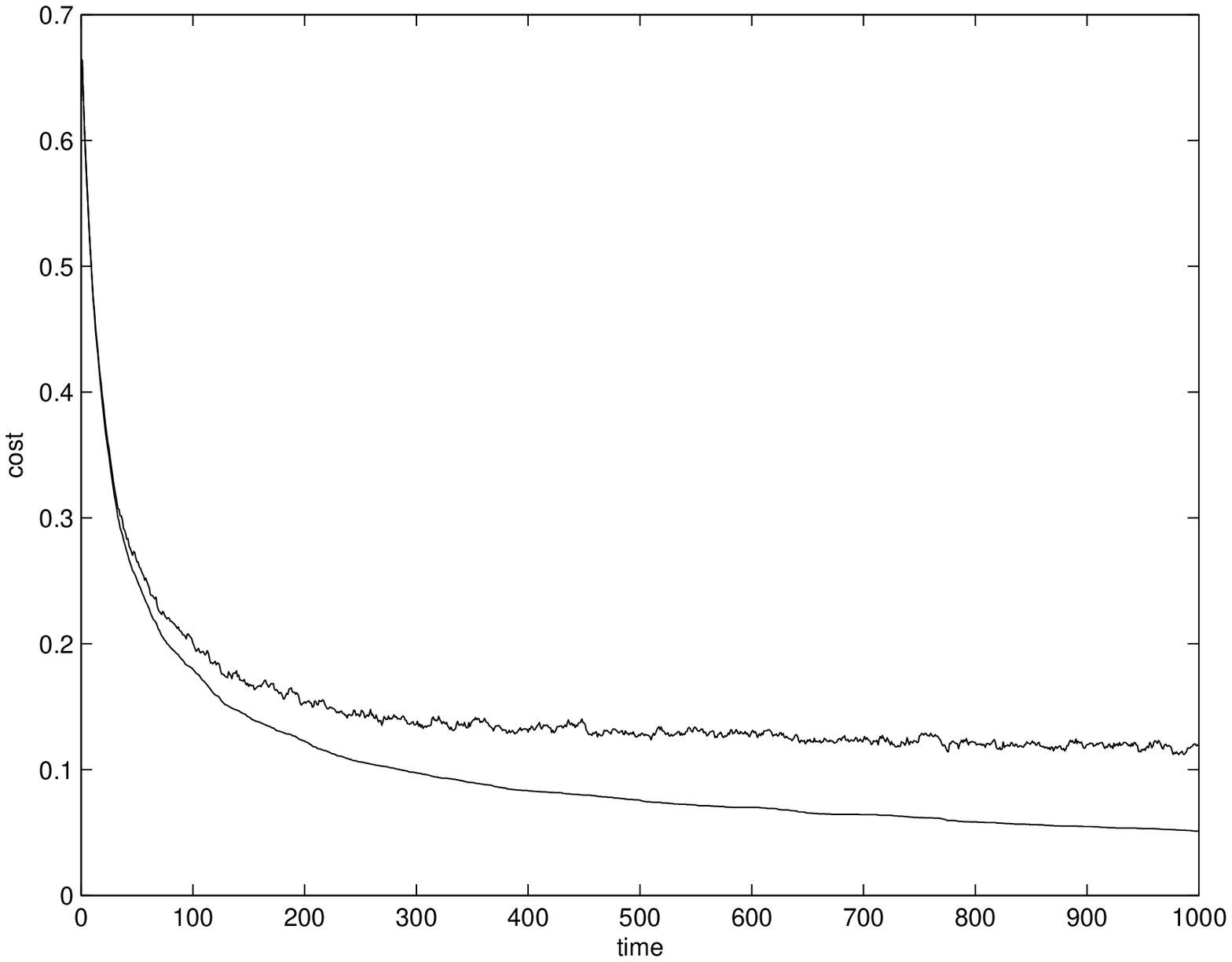}
\vspace{2mm}
\caption{Evolution of best and current cost as a function of time
for $\tau=1.5$-EO for an average over 10 $\gamma=5$-runs.}
\label{g5t15ave}
\end{figure}

The experiments show that optimization of
the approximate cost-matrices
using $\tau$-EO
gives a reasonably good, but not perfect, approximation to the 
perfect solution. More detailed study of
the results of tracking based on this method
need to be performed before the method can
be evaluated.

It is necessary to make much more tests to see which approximation
of the true cost matrix should be used.
Here we chose to use a matrix that is as close
to the phase transition as possible. The motivation
for this was that the problems should definitively be solvable.
By construction, this means that the average properties of
the cost matrix obtained from the sensor networks should have
$\gamma < \gamma_c$. However, it is not completely clear that
this should be valid for a specific instance of cost matrix
from a specific instance of scenario. This thus needs to 
be confirmed in future work.

\section{Discussion and future work}

In conclusion, this paper used
extremal optimization to do
pre-processing of sensor
reports before they are used by
single-target trackers. 
When receiving large bursts of
sensor reports from sensor networks
or swarming UAVs, traditional methods
for clustering and associating may
be too slow. The approach presented in this
paper has two components:

\begin{enumerate}
\item Since it costs to much to calculate the
complete $N \times N$ cost matrix when
$N$ is large and the cost of placing
two reports in the same cluster
is complicated, it is beneficial
to use an approximate cost matrix.
By taking advantage of results
on phase transitions occurring
for clustering problems
on random graphs, it is possible
to determine how many pairs of
conflict must be calculated in
order to get a fair approximation
of the full cost matrix.
\item We used the extremal optimization
algorithm to solve the approximate
problem.
\end{enumerate}

The method presented here can be extended
in a number of ways. The extremal
optimization method has been extended
to punish variables that are flipped
ofter~\cite{middleton}. This
leads to more efficient
exploration of
rugged energy landscapes. 
It would be
interesting to see how this affects the
run-time and quality of solutions
for the pre-processing method
presented here.

This dynamic clustering briefly mentioned
in the Introduction
was not
implemented in the experiments described here,
but the only changes needed in the algorithm
are that the $\lambda_i$ need to be re-calculated
taking into account also the new reports. Since
they have to be re-calculated in every time-step,
the dynamic version of the method has no
extra overhead compared to the static.
 
The extremal optimization method
is a general-purpose algorithm.
Among its advantages is that it
is an any-time algorithm, {\em i.e.},
it always has a ``best so far'' solution
that can be output and used. It would
be interesting to investigate
the behavior of extremal optimization
for resource allocation 
and scheduling problems.
In particular, extremal optimization
should be a better alternative than
genetic algorithms for many 
fusion applications.

The method used to determine the
approximate cost matrix is the simplest
possible --- it relies only on the most
basic information on the problem and
on phase transitions in the $\mathcal{G}(N,M)$
ensemble. By using more information
on the structure of graphs near the
phase transition, it might be possible
to get better approximations.
Another approach would be to try to 
capture the characteristics of
the complete cost matrix for specific
configurations of sensors and
use this information to get a better
approximation.

We caution the reader that while $c_{ij}$
seems to be a good approximation of $C(i,j)$ for the
problems tested in this paper, this might not
always be the case, and more thorough investigations
of this need to be done using real data. 
It might also be possible to use more
characteristics of the problem to construct
the approximate cost-matrix $c_{ij}$.
Examples of this could be to use
more advanced random graph models than
$\mathcal{G}(N,M)$, or dynamically
obtained information on the configuration
of our sensors or on the last known
formation of the enemy objects.

In order to extend the method
in the ways outlined above,
it should be tested using
sensor data from large-scale
fusion demonstrator systems,
such as IFD03~\cite{ifd03}.

{\small
\bibliographystyle{spiebib}
\bibliography{ref}
}
\end{document}